\documentclass[letterpaper]{article} % DO NOT CHANGE THIS
\usepackage{aaai24}  % DO NOT CHANGE THIS
\usepackage{times}  % DO NOT CHANGE THIS
\usepackage{helvet}  % DO NOT CHANGE THIS
\usepackage{courier}  % DO NOT CHANGE THIS
\usepackage[hyphens]{url}  % DO NOT CHANGE THIS
\usepackage{graphicx} % DO NOT CHANGE THIS
\usepackage{natbib}  % DO NOT CHANGE THIS AND DO NOT ADD ANY OPTIONS TO IT
\usepackage{caption} % DO NOT CHANGE THIS AND DO NOT ADD ANY OPTIONS TO IT
\usepackage{algorithm}

\usepackage{multirow}
\usepackage{paralist}
\usepackage{subfigure}
\usepackage{booktabs}
\usepackage[noend]{algpseudocode}
\usepackage{bbding}
\usepackage{newfloat}
\usepackage{listings}
\DeclareCaptionStyle{ruled}{labelfont=normalfont,labelsep=colon,strut=off} % DO NOT CHANGE THIS
\floatstyle{ruled}
\newfloat{listing}{tb}{lst}{}
\floatname{listing}{Listing}
\title{Panoptic Scene Graph Generation with Semantics-Prototype Learning}
\author{
    Li Li\textsuperscript{\rm 1},
    Wei Ji\textsuperscript{\rm 1}\textsuperscript{*},
    Yiming Wu\textsuperscript{\rm 2},
    Mengze Li\textsuperscript{\rm 3}\thanks{Corresponding Author.},
    You Qin\textsuperscript{\rm 1},
    Lina Wei\textsuperscript{\rm 4},
    Roger Zimmermann\textsuperscript{\rm 1}
}
\affiliations{
    \textsuperscript{\rm 1}National University of Singapore \\
    \textsuperscript{\rm 2}The University of Sydney \\
    \textsuperscript{\rm 3}Zhejiang University \\
    \textsuperscript{\rm 4}Hangzhou City University
    lili02@u.nus.edu,
    weiji0523@gmail.com,
    mengzeli@zju.edu.cn,
    e0962995@u.nus.edu
}

\usepackage{bibentry}
\begin{document}

\maketitle

\begin{abstract}
Panoptic Scene Graph Generation (PSG) parses objects and predicts their relationships (predicate) to connect human language and visual scenes.
However, different language preferences of annotators and semantic overlaps between predicates lead to biased predicate annotations in the dataset, i.e. different predicates for the same object pairs.
Biased predicate annotations make PSG models struggle in constructing a clear decision plane among predicates, which greatly hinders the real application of PSG models.
To address the intrinsic bias above, we propose a novel framework named ADTrans to adaptively transfer biased predicate annotations to informative and unified ones. To promise consistency and accuracy during the transfer process, we propose to observe the invariance degree of representations in each predicate class, and learn unbiased prototypes of predicates with different intensities. Meanwhile, we continuously measure the distribution changes between each presentation and its prototype, and constantly screen potentially biased data. Finally, with the unbiased predicate-prototype representation embedding space, biased annotations are easily identified.
Experiments show that ADTrans significantly improves the performance of benchmark models, achieving a new state-of-the-art performance, and shows great generalization and effectiveness on multiple datasets. Our code is released at \url{https://github.com/lili0415/PSG-biased-annotation}.
\end{abstract}

\section{Introduction}
Panoptic Scene Graph Generation (PSG) \cite{yang2022psg} aims to simultaneously detect instances and their relationships within visual scenes \cite{Chang_2023}. Instead of coarse bounding boxes used in Scene Graph Generation (SGG) \cite{10.1145/3474085.3475540,Xu_2017_CVPR,Lin_2020_CVPR,Li_2021_CVPR,pmlr-v119-chen20j,yao2022pevl,https://doi.org/10.48550/arxiv.2109.11797}, PSG proposed to construct more comprehensive scene graphs with panoptic segmentation \cite{Kirillov_2019_CVPR}. %\ul{This approach has} 
PSG methods have the potential to bridge the gap between visual scenes and human languages and thus has the ability to contribute to related vision-language tasks, such as image retrieval \cite{Johnson_2015_CVPR,wang2022global,wang2022recognition,lv2023duet}, image captioning \cite{Chen_2020_CVPR,Wei_Chen_Ji_Yue_Chua_2022}, and visual question answering \cite{Teney_2017_CVPR,Li_2022_CVPR,Antol_2015_ICCV,xiao2021video,fang2023you}.

%\ym{use vectorgraphs}
\begin{figure}[t]
\includegraphics[width=0.47\textwidth]{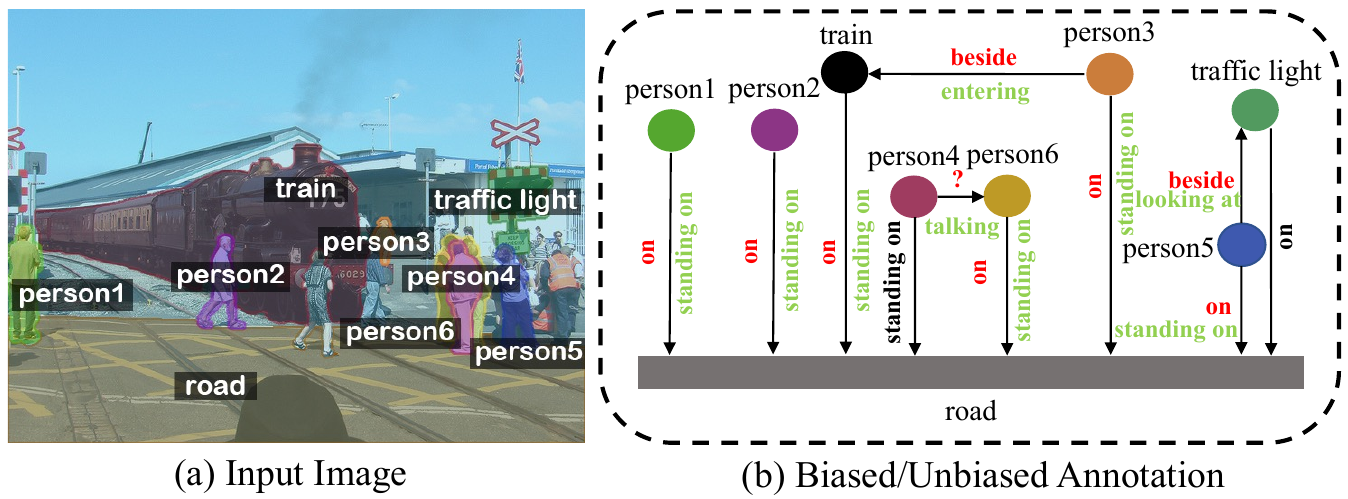} 
   \caption{(a) Exemplar panoptic segmentation results of an input image. (b) present annotation transfer process. Our proposed method promotes the original dataset (annotations in red) by identifying biased annotation and potentially positive samples, and then adaptively and accurately transferring them to target triplet pairs (annotation in green). }
   \label{fig:method}

\end{figure}

However, %\ul{the current performance of PSG methods is suboptimal due to the biased annotation problem, with generated scene graphs often containing biased and noisy information} 
PSG methods currently suffer from suboptimal performance due to biased and noisy information in generated scene graphs, stemming from the problem of \emph{biased annotations}. 

Exploring the inference mechanism of PSG and SGG models, it is translating visual scenes to linguistic descriptions, i.e., mapping visual instances to subjects/objects, and their relationships to predicates. We regard the problem above as the semantic ambiguity between predicates, and the contradictory mappings from visual to linguistics. 
There are a lot of semantic overlaps and hierarchical relationships among predicates, e.g., the superclass predicate \textit{on} for its subclass predicate \textit{standing on}. Because of the semantic overlaps and the inconsistent preferences of annotators, contradictory mappings from visual to linguistics unavoidably exist in the training dataset, and deteriorate the long-tail distribution problem of the training dataset. As shown in Fig.~\ref{fig:method}, with the difficulty of applying a unified standard for annotation, annotators tend to annotate general predicate labels (e.g., “on” and “beside”) for simplicity instead of informative ones (e.g., “standing on” and “looking at”). As a result, models cannot learn a consistent mapping from visual to semantics, instead, they entangle predicate labels and the prior knowledge of long-tail distribution, leading to serious harm for model training stage.

Previous works \cite{Zellers_2018_CVPR,Yu2020CogTreeCT,yao2022pevl,Xu_2017_CVPR,Tang_2019_CVPR,Tang_2020_CVPR,Lin_2020_CVPR} exploit numerous model architectures to alleviate the bias problem, but these models trained by biased datasets achieve relatively limited performances, and cannot fundamentally solve the problem. \citet{zhang2022fine} have proposed to enhance the training dataset by a data transfer framework, which transfers head predicate labeled samples to tail predicate labeled ones. However, their framework inaccurately transfers a significant number of samples, leading to imbalanced performance among predicates.

%However, their method keeps a fixed ratio during data transfer process, leading to a huge number of wrongly transferred samples. The results show that the model trained using the dataset enhanced by their method shows a substantial reduction in the recall rate of head predicate labels.

To alleviate the biased annotation problem, we propose to construct a promised and reasonable dataset, which includes plentiful samples with consistent predicate annotation. Specifically, there exist two types of predicates that need to be refined: indistinguishable triplet pairs with semantic ambiguity and potentially positive samples missed by annotators \cite{zhang2022fine}. We transfer these two kinds of biased annotations to high-quality consistent predicate annotations \cite{zhang2023revisiting,lv2022personalizing}. 

We introduce a new adaptive data transfer framework named ADTrans for PSG. Our framework emphasizes consistency during the data transfer process, and performs the data transfer process adaptively and accurately. Besides the prior knowledge of dataset distribution, we believe textual information alignment helps in building consistency during the data transfer process. A general way is leveraging language models to extract semantic embeddings, however, words embedding vectors generated by language models often have high similarity because of the broad class intersection of the open world, leading to the misalignment of the textual domain and the relationship domain. Thus, we propose a prototype-based predicate representation learning method. Unbiased predicate representations are expected to share invariant features within each predicate class. Thus, we employ contrastive learning to increase intra-class cohesion and inter-class separation, while focusing more on hard samples (visually similar predicates). Meanwhile, we observe the invariance degree of representations in each predicate class, and learn predicate prototypes with dynamic intensities. We continuously measure the distribution changes between each presentation and its prototype, and constantly screen potentially biased data. Finally, with the unbiased predicate representation embedding space, biased annotations are easily identified and transferred.

In summary, the following contributions are made in this paper: 
\begin{itemize}
    \item A novel, plug-and-play framework named ADTrans is proposed, which aims at adaptively and accurately performing data transfer to promise a reasonable dataset with informative and standard-unified labels, and more solid training samples. 
    \item We propose a new prototype-based predicate representation learning method, aiming at a reasonable information alignment process between the textual domain and the relationship domain, to promise consistency during the data transfer process.
    \item Comprehensive experiments demonstrate that the proposed method shows validity on two datasets, and significantly enhances the performance of benchmark models, achZieving new state-of-the-art performances.
\end{itemize}

\section{Related Work}
\noindent\textbf{Panoptic Scene Graph Generation.} SGG has gained increasing attention from the computer vision community for its promising future in high-level vision-language tasks \cite{Teney_2017_CVPR,Li_2022_CVPR,Antol_2015_ICCV,Chen_2020_CVPR,Johnson_2015_CVPR}. Early two-stage methods divide the whole task into objects locating process and relationships prediction process, and struggle for a better feature extraction network \cite{Xu_2017_CVPR,Zellers_2018_CVPR,Tang_2019_CVPR,Lin_2020_CVPR}. More recently, a novel task named Panoptic Scene Graph Generation (PSG) \cite{yang2022psg}, which points out that it will contain noisy and ambiguous pixels if only coarse bounding boxes are provided, and aims at constructing more comprehensive scene graphs with panoptic segmentation rather than coarse bounding boxes. In addition, the provided ground truth panoptic segmentation can also significantly promote the performance of even the most classic SGG method, IMP \cite{Xu_2017_CVPR}.

\noindent\textbf{Towards Debiasing Scene Graph Generation.} 
\cite{Zellers_2018_CVPR} first introduces the biased prediction problem in SGG. \cite{Tang_2019_CVPR} and \cite{Chen_2019_CVPR} provide a more reasonable metric (Mean Recall) aiming at calculating the recall of each predicate label independently. To directly face the problem, causal inference framework \cite{Tang_2020_CVPR,Zhang_2022_ACCV} is applied to alleviate data bias during the inference process, and CogTree \cite{Yu2020CogTreeCT} is designed to train models with the ability to make informative predictions on predicate labels. More recently, \cite{zhang2022fine,LiMM} argues that performance could be promoted if there is a reasonable and sound dataset. However, the data transfer method in \cite{zhang2022fine} is efficient but so rigid and inflexible that a huge number of positive samples will be wrongly transferred due to its simple and coarse transfer method, which will lead to a remarkable decline in recall rate of head predicate labels. In this paper, we provide a new approach for SGG and PSG datasets debiasing.

\begin{figure*}
\includegraphics[width=0.99\textwidth]{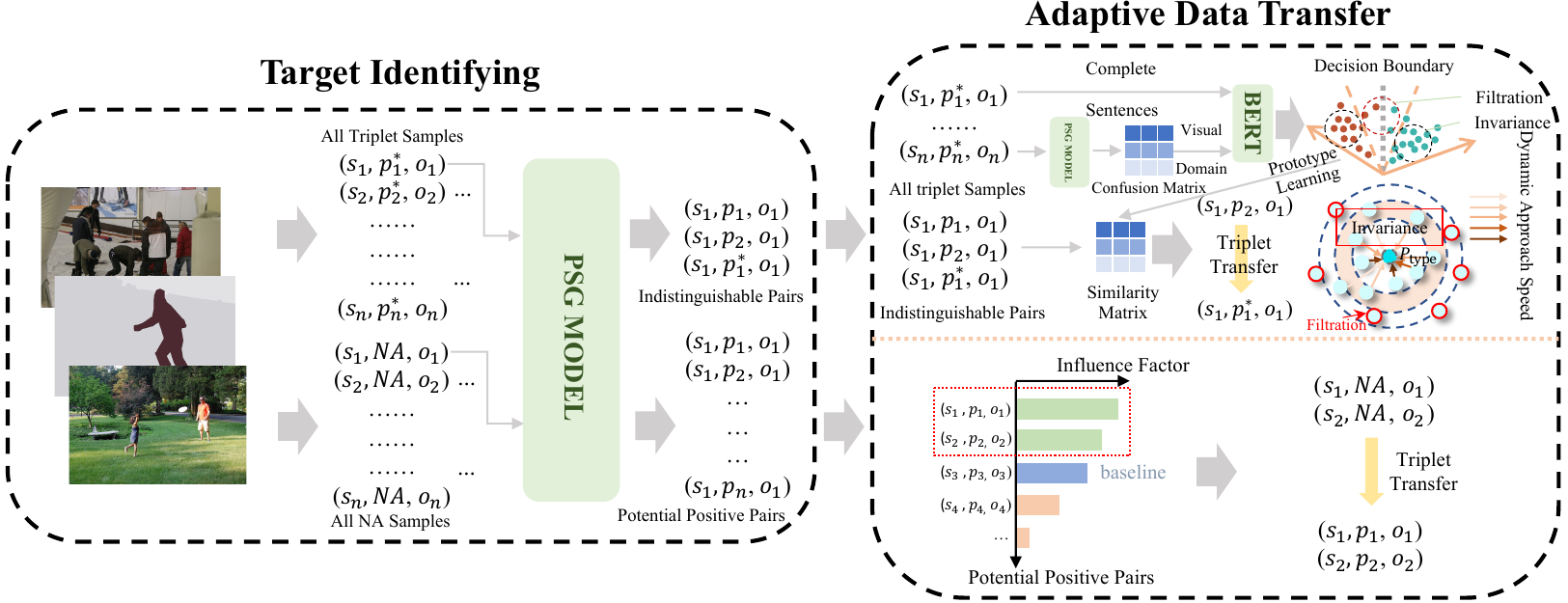} 
 \caption{Illustration of the overall pipeline. It learns unbiased semantics-prototypes and the learned prototypes help to promise the consistency during data transfer process.}
 \label{model structure}

\end{figure*}

\section{Method}
In this section, we first introduce the biased annotation identifying method. Then we introduce relation representation extraction. After that, we provide the detailed semantics-prototype learning method. Finally, the data transfer method and resampling method are introduced.

\noindent{\textbf{Target Identifying.}} Following \citet{zhang2022fine}, we identify indistinguishable triplet pairs by checking the inconsistency between the model's predictions and the ground truth labels. To be more specific, we use a pre-trained model (e.g. VCTree\cite{Tang_2019_CVPR}) to predict predicate labels for every pair of ground truth subject and object pairs in the PSG training dataset and identify possible indistinguishable predicate labels.
For potentially positive samples, we also use a pre-trained model to predict predicate labels for every pair of ground truth subject and object labels that have not yet been annotated with predicate labels, also known as NA samples.

\subsection{Relation Representation Extraction}
\label{RRE}
To make the language model more sensitive to predicate semantics, we fine-tune the language model with contrastive relation representation training \cite{li2022hero}.

\noindent \textbf{Robust Contrastive Training.} We first collect all of the triplets that appear in the training set.
Each triplet will be converted to a sentence for language model processing.
For example, $\textless$ person, standing on, snow $\textgreater$ will be converted to \textit{The person is standing on the snow}.
% Among all of the sentences, we consider the sentences with the same predicates as the positive pair and sentences with different predicates as the negative pair.

Formally, given a sentence $s_i$ with predicate $p_i$ in the batch $S=\left\{s_k\right\}_{i=k}^N$, we can construct its positive set ${PS}_i=\{s_k|p_i=p_k \}_{k\neq i}$ and negative set ${PN}_i=\{s_k|p_i \neq p_k \}_{k\neq i}$.
With the training data, we use a InfoNCE loss to optimize the language model:

\begin{equation}
\label{eq:rct}
L_{l m}= 
\sum_{\mathrm{i}=1}^{\mathrm{N}}-\log \frac{f_{pos}}{f_{pos}+f_{neg}} ,
\end{equation}
where:
\begin{equation}
    f_{pos} = \sum_{s_{j} \in P S_{i}} e^{\sin \left(h_{i}, h_{j}\right) / \mathrm{T}},
\end{equation}
\begin{equation}
    f_{neg} = \sum_{s_{g} \in N G_{i}} e^{\sin \left(h_{i}, h_{g}\right) / \mathrm{T}}.
\end{equation}
$\mathrm{T}$ is a temperature hyper-parameter, \textit{N} is batch size, $h_{i,j,g}$ are language model generated sentence representations for $s_{i,j,g}$, and $sim\left(h_i,h_j\right)$ is the cosine similarity.

% \noindent \textbf{Angular Margin for Robust Training.}
To further boost the sensitivity to predicate similarity, we additionally introduce an angular margin $m$ for the positive pairs.
Formally, instead of using the previously defined $e^{sim ( h_{i},h_{j} )/T}$, we use:
\begin{equation}
e^{sim ( h_{i},h_{j} )/\mathrm{T}} \to e^{cos ( \theta _{i,j}+m )/\mathrm{T} },
\end{equation}
where $ \theta _{i,j} $ is the arc-cosine similarity between $ i $ and $ j $.
% \begin{equation}
% \theta_{i,j}=arccos\left(\frac{h_i^T\times h_j}{|\left|h_i\right||\times||h_j||}\right).
% \end{equation}
% \begin{equation}
% LPS={\textstyle \sum_{j\in PS_{i} }e^{cos\left ( \theta _{i,j}+m    \right )/T }}
% \end{equation}
Thus, $f_{pos}$ becomes:
\begin{equation}
    f_{pos} = \sum_{s_{j} \in P S_{i}} e^{cos ( \theta _{i,j}+m )/\mathrm{T}}.
\end{equation}

\noindent \textbf{Alignment with Visual Domain.} There is typically a domain gap between the visual domain and the textual domain, i.e., the textual similar predicates are not visually similar.
To align the language model with the visual domain \cite{li2023winner,fang2022multi}, we try to incorporate some visual prior knowledge into the language model fine-tuning.

Specifically, we take advantage of the confusion matrix $C \in \mathbf{R}^{Q\times Q}$ generated by a pre-trained VCTree model, where $ Q $ is the number of predicates in the dataset, and $C_{i,j}$ denotes the averaged prediction score for predicate $p_j$ on all examples annotated with $p_i$.
When $C_{i,j}$ is near to 1 (very high), $p_i$ and $p_j$ are visually similar and when $C_{i,j}$ is near to 0 (very low), $p_i$ and $p_j$ are visually different.

For the language model training, we expect the language model to be aligned with the visual domain similarity judgment.
Thus, the language model should also distinguish the visually different predicate pairs but avoid the visually similar predicate pairs from being too distant in the feature space.
The metric $1-C_{i,j}$ can satisfy our requirement.

Formally, we add $1-C_{i,j}$ as a weight for negative pairs:
\begin{equation}
    e^{cos\left ( \theta _{i,j}    \right )/\mathrm{T} } \to \left ( 1-C_{i,j}  \right ) \times e^{cos\left ( \theta _{i,j}    \right )/\mathrm{T} }. 
\end{equation}

Then, the $f_{neg}$ becomes:
\begin{equation}
    f_{neg} = \sum_{s_{g} \in PN_{i}} \left (1-C_{i,g}   \right )e^{\cos\left(\theta_{i,g}\right) / \mathrm{T}}.
\end{equation}

\noindent{\textbf{Invariant Representation Exploration.}} We propose the invariant representation exploration to further promise the unbiased representation of predicates.

Specifically, we say that an unbiased representation $ \Phi :X\to H $ elicits an invariant predictor $ \omega  $ across positive set $ \varepsilon  $ if there is an optimizer $ \omega :H\to Y $ simultaneously optimal for all samples from the positive set. The learning objective can be formulated as:

\begin{equation}
\label{eq:irmo}
    \omega \in argmin_{\bar{\omega}:H\to Y}O^{e}(\omega,\Phi).
\end{equation}

Eq.~\ref{eq:irmo} tries to learn a feature representation from $ \Phi (\cdot) $ that can induce an optimizer $ \omega(\cdot) $ which is simultaneously optimal for all $ e \in \varepsilon $.
Thus, we propose the invariant representation regularization which can be formulated as:

\begin{equation}
\label{eq:irm}
    L_{irm}=\sum_{i=1}^{N}\lambda Var(L^{i} ),
\end{equation}
where $ L^i=\left\{L_{lm}\left(j\right)|p_i=p_j\right\} $ denotes the loss values of samples from the positive set, and $ \lambda $ is a hyper-parameter. The minimization of variances of loss values encourages unbiased representation learning for each predicate class \cite{arjovsky2020invariant}.

% \begin{figure*}[!t]
% \includegraphics[width=6.85in,height=3.25in]{./images/loss.pdf} 
% \label{fig:sent_rep}
%  \caption{The framework of sentence representation learning module. We supplement triplet pairs to complete sentences, and then fine-tune a pre-trained language model using these sentences to enhances its discriminative power on indistinguishable predicates.}
% \vspace{-1.5em}
% \end{figure*}

\begin{table*}[]
\tabcolsep=0.08cm

\begin{tabular}{cr|ccccccccc}
\hline
\multicolumn{2}{c|}{\multirow{2}{*}{Method}}                                      & \multicolumn{9}{c}{Scene Graph Generation}                                                                                                                    \\ \cline{3-11} 
\multicolumn{2}{c|}{}                                                             & R@20 & R@50 & \multicolumn{1}{c|}{R@100} & mR@20         & mR@50         & \multicolumn{1}{c|}{mR@100}        & PR@20         & PR@50         & PR@100        \\ \hline
\multicolumn{1}{c|}{\multirow{12}{*}{Two-Stage}} & \multicolumn{1}{l|}{IMP \cite{Xu_2017_CVPR}}       & 16.5 & 18.2 & \multicolumn{1}{c|}{18.6}  & 6.52          & 7.05          & \multicolumn{1}{c|}{7.23}          & 12.9          & 13.7          & 13.9          \\
\multicolumn{1}{c|}{}                            &  \quad +IETrans \cite{zhang2022fine}                      & 14.5 & 15.9 & \multicolumn{1}{c|}{16.4}  & 10.2          & 11.0          & \multicolumn{1}{c|}{11.3}          & 14.5          & 15.4          & 15.7          \\
\multicolumn{1}{c|}{}                            &  \multicolumn{1}{l|}{\quad +ADTrans }                    & 15.0 & 16.5 & \multicolumn{1}{c|}{17.0}  & \textbf{12.5} & \textbf{13.5} & \multicolumn{1}{c|}{\textbf{14.0}} & \textbf{16.0} & \textbf{17.1} & \textbf{17.5} \\ \cline{2-11} 
\multicolumn{1}{c|}{}                            & \multicolumn{1}{l|}{VCTree \cite{Tang_2019_CVPR}}    & 20.6 & 22.1 & \multicolumn{1}{c|}{22.5}  & 9.70          & 10.2          & \multicolumn{1}{c|}{10.2}          & 16.0          & 16.8          & 16.9          \\
\multicolumn{1}{c|}{}                            & \quad +IETrans \cite{zhang2022fine}                      & 17.5 & 18.9 & \multicolumn{1}{c|}{19.3}  & 17.1          & 18.0          & \multicolumn{1}{c|}{18.1}          & 19.6          & 20.5          & 20.7          \\
\multicolumn{1}{c|}{}                            & \multicolumn{1}{l|}{\quad +ADTrans }                        & 17.9 & 19.5 & \multicolumn{1}{c|}{19.9}  & \textbf{18.0} & \textbf{18.9} & \multicolumn{1}{c|}{\textbf{19.0}} & \textbf{20.2} & \textbf{21.2} & \textbf{21.4} \\ \cline{2-11} 
\multicolumn{1}{c|}{}                            & \multicolumn{1}{l|}{MOTIFS \cite{Zellers_2018_CVPR}}    & 20.0 & 21.7 & \multicolumn{1}{c|}{22.0}  & 9.10          & 9.57          & \multicolumn{1}{c|}{9.69}          & 15.5          & 16.3          & 16.5          \\
\multicolumn{1}{c|}{}                            & \quad +IETrans \cite{zhang2022fine}                      & 16.7 & 18.3 & \multicolumn{1}{c|}{18.8}  & 15.3          & 16.5          & \multicolumn{1}{c|}{16.7}          & 18.2          & 19.4          & 19.7          \\
\multicolumn{1}{c|}{}                            & \multicolumn{1}{l|}{\quad +ADTrans }                        & 17.1 & 18.6 & \multicolumn{1}{c|}{19.0}  & \textbf{17.1} & \textbf{18.0} & \multicolumn{1}{c|}{\textbf{18.5}} & \textbf{19.4} & \textbf{20.4} & \textbf{20.8} \\ \cline{2-11} 
\multicolumn{1}{c|}{}                            & \multicolumn{1}{l|}{GPSnet \cite{Lin_2020_CVPR}}    & 17.8 & 19.6 & \multicolumn{1}{c|}{20.1}  & 7.03          & 7.49          & \multicolumn{1}{c|}{7.67}          & 13.6          & 14.4          & 14.7          \\
\multicolumn{1}{c|}{}                            & \quad +IETrans \cite{zhang2022fine}                      & 14.6 & 16.0 & \multicolumn{1}{c|}{16.7}  & 11.5          & 12.3          & \multicolumn{1}{c|}{12.4}          & 15.3          & 16.2          & 16.5          \\
\multicolumn{1}{c|}{}                            & \multicolumn{1}{l|}{\quad +ADTrans }                        & 17.8 & 19.2 & \multicolumn{1}{c|}{19.5}  & \textbf{16.5} & \textbf{17.5} & \multicolumn{1}{c|}{\textbf{17.6}} & \textbf{19.3} & \textbf{20.3} & \textbf{20.5} \\ \hline
\multicolumn{1}{c|}{\multirow{3}{*}{One-Stage}}  
& \multicolumn{1}{l|}{PSGTR \cite{yang2022psg}}     & 28.4 & 34.4 & \multicolumn{1}{c|}{36.3}  & 16.6          & 20.8          & \multicolumn{1}{c|}{22.1}          & 21.9          & 26.3          & 27.6          \\
\multicolumn{1}{c|}{}                            & \quad +IETrans \cite{zhang2022fine}                      & 25.3 & 28.8 & \multicolumn{1}{c|}{29.2}  & 23.1          & 27.2          & \multicolumn{1}{c|}{27.5}          & 24.9          & 28.4          & 28.7          \\
\multicolumn{1}{c|}{}                            & \multicolumn{1}{l|}{\quad +ADTrans }                          & 26.0 & 29.6 & \multicolumn{1}{c|}{30.0}  & \textbf{26.4} & \textbf{29.7} & \multicolumn{1}{c|}{\textbf{30.0}} & \textbf{27.1} & \textbf{30.2} & \textbf{30.5} \\ \hline
\end{tabular}
\label{main}
\caption{The results (R@K, mR@K and PR@K) on SGDet task of our method and other baselines on PSG dataset. IETrans and ADTrans denote models equipped with different dataset-enhancement methods. } 
\end{table*}

\subsection{Semantics-prototype Learning}
\label{SPL}
The extracted relation representation may be suffered from possibly biased annotations. Thus, we utilize dynamic prototype learning to maximize the discriminative power between predicate classes. We measure the invariance within each predicate class, and discard biased data by stage. This promise unbiased predicate representation embedding space for accurate data transfer.

\noindent{\textbf{Dynamic Prototype Updating.}} To build an unbiased embedding space for predicates, we further propose to construct the prototype space for predicates. Specifically, we specify the total prototype space $ P_{type} \in \mathbf{R}^{L \times Q} $, where $ L $ is the same as the size of semantic embedding. Then we dynamically update the prototype space depending on the degree of invariance during the robust contrastive training process.

Given a batch $ S $, we can construct multiple positive sets $ PS=\left \{ ps_1,ps_2,...,ps_P \right \}  $ with different predicates, where $ P $ is the number of different predicates in the batch $ S $.

For every positive set in the $ PS $ with predicate $ p_i $, we obtain its average predicate representation embedding as follows:

\begin{equation}
    H_{aver}^{p_i}=\frac{1}{N^{p_i}} {\textstyle \sum_{s=1}^{N^{p_i}}} h_i, 
\end{equation}
where $ N^{p_i} $ denotes the number of samples with predicate $ p_i $ in the batch $ S $, and $ h_i $ denotes the predicate representation embedding. We average the summation of all predicate representations with the same predicate in the batch to get the average feature embedding.

With the help of observed invariant representation, we update the prototype space for predicate $ p_i $ with a moving average approach:

\begin{equation}
    % P_{type}^{p_i}=\beta P_{type}^{p_i}+(1-\beta) \frac{1}{ \gamma Var(L^{i})N^{p_i} } (F_{aver}^{p_i}-P_{type}^{p_i}),
    P_{type}^{p_i}=\beta P_{type}^{p_i}+(1-\beta) \underbrace{\frac{1}{ \gamma Var(L^{i})N^{p_i} }}_{Approach\;Speed} (H_{aver}^{p_i}-P_{type}^{p_i}),
\end{equation}
where $ \beta $ and $ \gamma $ are hyper-parameters.

\noindent{\textbf{Multistage Data Filtration.}}
\label{MDC}
Biased and noisy samples in the training dataset are certain to influence the unbiased predicate representation learning process. Thus, We design the multistage data filtration to multistage-ly filter out these bad samples. Specifically, we take advantage of the invariant representation regularization and the sample-prototype distribution shift as the measurements for the sample's quality. 

\begin{figure}

\includegraphics[width=0.47\textwidth]{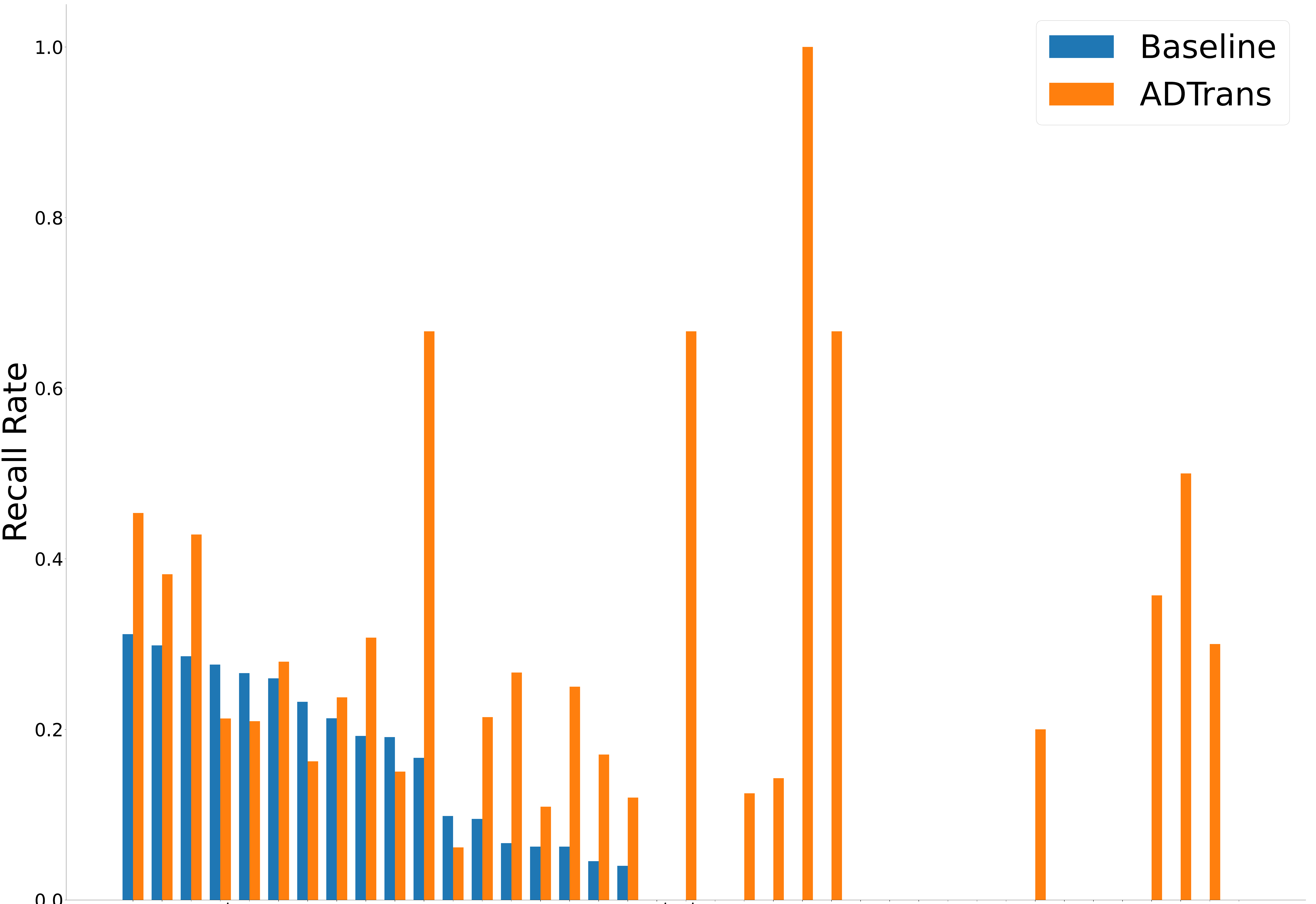} 
 \caption{R@100 for predicates under SGDet task among plain PSGTR, and PSGTR with ADTrans. ADTrans achieves more balanced and effective predicate discrimination among predicates with different frequencies than plain PSGTR (The horizontal axis, moving from left to right, illustrates predicates arranged in order of high frequency to low frequency).}
 \label{compare}
\end{figure}

For every training epoch, we collect $ V\in\mathbf{R}^{G} $ from Eq.~\ref{eq:irm}, which denotes the variance of loss values of every training sample in the training dataset with $ G $ samples. Then we average the collected variances on predicate labels, getting $ V_{aver}\in\mathbf{R}^{Q} $. For every sample $ S_i $ with predicate label $ p_i $ and variance $ V_i $ in the training dataset, we judge whether it is part of potentially biased and noisy samples, which can be formulated as:

\begin{equation}
    P_{bn}=\left \{ S_i | V_i>\mu V_{aver}^{i}  (H_{aver}^{p_i}-P_{type}^{p_i})  \right \}, 
\end{equation}

where $ V_{aver}^{i} $ denotes the averaged variance on predicate label $ p_i $, and $ \mu $ is a hyper-parameter. We further sort $ P_{bn} $ by the loss value derived from Eq.~\ref{eq:rct} and drop out the top $ D\% $ of training data. If there are fewer than 100 samples in a predicate class, we do not drop out any more samples from it.

The multistage data filtration avoids the influence of a large number of biased annotations (outlier noise). Thus, the whole unbiased predicate representation learning process promises the unbiased representation of predicates.

\subsection{Data Transfer}
\label{DT}
As a result, a similarity matrix $S\in\mathbf{R}^{Q\times Q}$ can be generated by calculating the cosine similarities between all prototypes.

For indistinguishable triplets, we directly use the similarity score as an adaptive transfer ratio. 

For potentially positive samples, we further define an influence factor, where the intuition is to transfer more data for the scarce relation triplets with low NA scores.

Formally, the influence factor is: 
\begin{equation}
{E}_{\left(s_i,p_i,o_i\right)}=\sqrt{-log\left({NA}_{score}\right)\times{\rm c}_{\left(si,oi\right)}\times{c}_{p_i}},
\end{equation}
where $-log(NA_{score})$ is the NA score, $c(s_i, o_i)$ is the scarcity of triplets with subject $s_i$ and $o_i$, and $ c_{p_i} $ is the scarcity of triplets with predicate $ p_i $.

% \begin{equation}
% {c}_{\left(s_i,o_i\right)}\ =\ \sqrt{\frac{1}{\sum_{p\in P} I_{\left(s_i,p,o_i\right)}}},
% \end{equation}
% and $c_{p_i}$ is:
% \begin{equation}
% {c}_{p_i}\ =\ \sqrt{\frac{1}{\sum_{s,o\in C} I_{\left(s,p_i,o\right)}}}.
% \end{equation}

In practice, we further normalize $c_{(s_i, o_i)}$ with softmax.

To judge whether a NA sample should be transferred, we rank all potential target triplet pairs according to their influence factors, and conduct the transfer of top $K_g\%$ pairs.

\subsection{Resampling}
\label{Re}
Without conflicts, we can directly integrate the transferred indistinguishable pairs and potentially positive pairs. Furthermore, a special re-sampling method is introduced on the integrated dataset to enhance it further.
We propose a new repeat factor for the task. For every triplet $(s_i,p_i,o_i)$ in each image, we calculate its repeat factor as:

\begin{equation}
{R}=max\left(1,t\times{\rm c}_{\left(si,oi\right)}\times{c}_{\left(p_i\right)}\right),
\end{equation}
where \textit{t} is a hyper-parameter controlling the possible repeat times. The maximum value of the repeat factor within each image is then selected.

\section{Experiment}

\subsection{Dataset and Evaluation Metrics}
\noindent\textbf{Dataset}. We evaluate our method on Visual Genome \cite{Krishna2017} and PSG dataset \cite{yang2022psg}.

\begin{figure*}[!t]
\centering   \includegraphics[width=\linewidth,height=5cm]{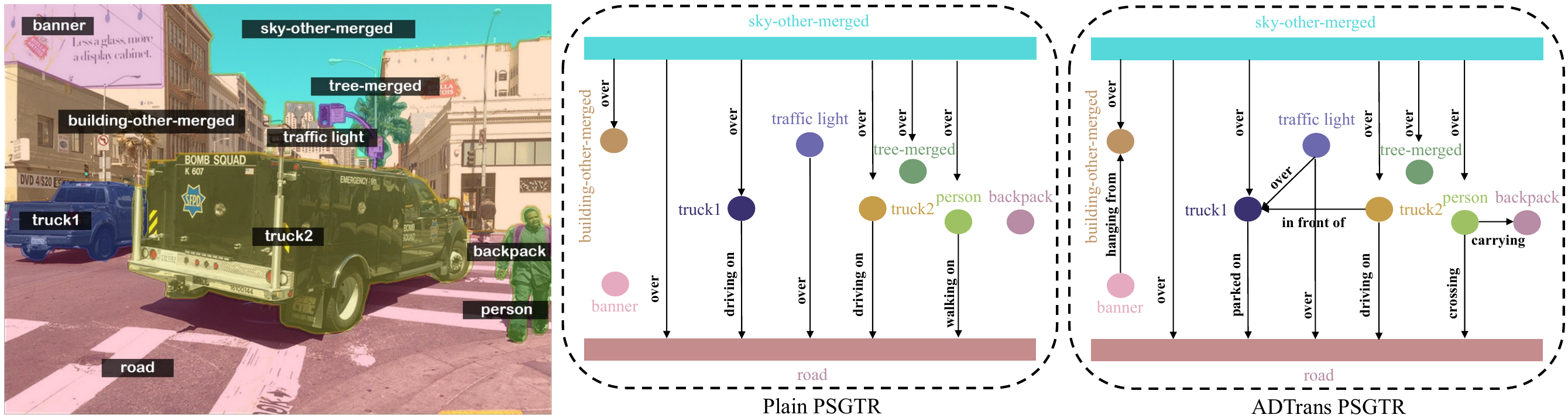}
    \caption{Visualization of plain PSGTR model and PSGTR equipped with our ADTrans. PSGTR with ADTrans can predict relationships between instances with greater accuracy and also select predicates that better match the visual scene.}
    \label{fig:performance}

\end{figure*}

% PSG dataset \cite{yang2022psg} has 47874 valid samples (45697 samples for training and 2177 samples for validation) with 56 predicate classes, and 80 thing and 53 stuff classes. For each image in the PSG dataset, there are an average of 11.0 instances and 5.6 relationships. Besides, we also evaluate our method on Visual Genome dataset. The Visual Genome dataset \cite{Krishna2017} comprises a total of 108,073 images. For this study, we employed the widely-used data splits outlined in the work of \cite{Xu_2017_CVPR}, which retain the 150 most frequently occurring object categories and 50 most commonly occurring predicate categories. The dataset is divided into a training set consisting of 70\% of the images, and a test set comprised of the remaining 30\%. To create a validation set, we followed the approach presented in the work of \cite{Zellers_2018_CVPR}, where a random sample of 5,000 images is taken from the training set.

\noindent\textbf{Evaluation Metric.} Following previous works \cite{Tang_2020_CVPR,Zellers_2018_CVPR}, we take recall@K (R@K) and mean recall@K (mR@K)\cite{Tang_2019_CVPR,Chen_2019_CVPR} as evaluation metrics. We also adopt a new evaluation metrics named percentile recall (PR), which can be formulated as $PR=30\%R+60\%mR+10\%PQ$, where PQ measures the quality of a predicted panoptic segmentation relative to the ground truth \cite{Kirillov_2019_CVPR}. For conventional SGG tasks on the VG dataset, we also adopt an overall metric F@K \cite{zhang2022fine}, which is the harmonic average of R@K and mR@K.

\begin{table}[]
\tabcolsep=0.1cm

\begin{tabular}{l|cccccc}
\hline
\multirow{2}{*}{Method} & \multicolumn{6}{c}{Predicate Classification}                                                                      \\ \cline{2-7} 
                        & mR@20         & @50         & \multicolumn{1}{c|}{@100}        & F@20          & @50          & @100         \\ \hline
MOTIFS                  & 11.7          & 15.2          & \multicolumn{1}{c|}{16.2}          & 19.5          & 24.5          & 26.0          \\
\quad+ADTrans                & \textbf{29.0} & \textbf{36.2} & \multicolumn{1}{c|}{\textbf{38.8}} & \textbf{36.1} & \textbf{41.7} & \textbf{43.5} \\ \hline
VCTree                  & 14.0          & 16.3          & \multicolumn{1}{c|}{17.7}          & 22.7          & 26.0          & 28.0          \\
\quad+ADTrans                & \textbf{30.0} & \textbf{32.9} & \multicolumn{1}{c|}{\textbf{35.5}} & \textbf{37.2} & \textbf{40.5} & \textbf{42.5} \\ \hline
GPSnet                  & 13.2          & 15.0          & \multicolumn{1}{c|}{16.0}          & 21.7          & 24.4          & 25.8          \\
\quad+ADTrans                & \textbf{27.3} & \textbf{32.0} & \multicolumn{1}{c|}{\textbf{34.7}} & \textbf{34.8} & \textbf{40.2} & \textbf{42.1} \\ \hline
\end{tabular}
\caption{The results (mR@K and F@K) on PREDCLS task of our method and other baselines on VG dataset.}
\label{predcls}
\end{table}

\subsection{Tasks and Implementation Details}

\noindent{\textbf{Tasks.}} We evaluate our method on three classic SGG tasks: Predicate Classification (PREDCLS), Scene Graph Classification (SGCLS) and Scene Graph Generation (SGDET).

\noindent\textbf{Implementation Details}. For the pre-trained language model, we use pre-trained {BERT-base} \cite{devlin-etal-2019-bert}. The decision margin \textit{m} is set to 10 degrees, the temperature hyper-parameter T is set to 0.05, and we use an AdamW \cite{Loshchilov} optimizer with a learning rate 2e-5. The hyper-parameter $ \lambda $ is set to 0.3, $ \beta $ is set to 5e5, and $ \gamma $ is set to 1.5.  The $ D $ in multistage data filtration is set to 50. For NA sample transfer, the value of $K_g$ is set to 0.05, and for the re-sampling process, the value of $t$ is set to 3e7.

\subsection{Qualitative Analysis}
As shown in Figure~\ref{fig:performance}, we can compare results predicted by plain PSGTR and PSGTR equipped with our ADTrans. Obviously, PSGTR with ADTrans can predict more accurate relationships between instances and also predict predicate that better fits the scene. 
% For instance, plain PSGTR predicts ``walking on" for the person on the right, and PSGTR with ADTrans predicts ``crossing", which fits the given scene better. 
We believe our method helps construct a more comprehensive scene graph.

\subsection{Comparison with State-of-the-Art Methods}
In this section, we report the results for ADTrans on different datasets, tasks, and baseline methods. All models use ResNet-50 \cite{He_2016_CVPR} as their backbones.
% As shown in Table.~\ref{main} and Table.~\ref{psgpredcls}, we report the results of our ADTrans method and baselines with different backbones for PSG \cite{yang2022psg} on predicate classification and scene graph generation tasks.

% Besides, as shown in Table.~\ref{sgg}, Table.~\ref{predcls} and Table.~\ref{sgcls}, we also report the results of our ADTrans method and baselines for conventional SGG on predicate classification, scene graph classification and scene graph generation tasks.

\begin{table}[]
\tabcolsep=0.1cm

\begin{tabular}{l|cccccc}
\hline
\multirow{2}{*}{Method} & \multicolumn{6}{c}{Scene Graph Classification}                                                                     \\ \cline{2-7} 
                        & mR@20         & @50         & \multicolumn{1}{c|}{@100}        & F@20          & @50          & @100         \\ \hline
MOTIFS                  & 6.0           & 8.0           & \multicolumn{1}{c|}{8.5}           & 10.1          & 13.1          & 13.8          \\
\quad+ADTrans                & \textbf{14.8} & \textbf{17.0} & \multicolumn{1}{c|}{\textbf{17.8}} & \textbf{20.2} & \textbf{22.5} & \textbf{23.7} \\ \hline
VCTree                  & 6.3           & 7.5           & \multicolumn{1}{c|}{8.0}           & 10.7          & 12.5          & 13.3          \\
\quad+ADTrans                & \textbf{16.0} & \textbf{19.0} & \multicolumn{1}{c|}{\textbf{19.8}} & \textbf{20.3} & \textbf{23.7} & \textbf{24.5} \\ \hline
GPSnet                  & 10.0          & 11.8          & \multicolumn{1}{c|}{12.6}          & 15.7          & 17.9          & 18.9          \\
\quad+ADTrans                & \textbf{15.5} & \textbf{18.2} & \multicolumn{1}{c|}{\textbf{18.8}} & \textbf{19.9} & \textbf{22.5} & \textbf{23.7} \\ \hline
\end{tabular}
\caption{The results (mR@K and F@K) on SGCLS task of our method and other baselines on VG dataset.}
\label{sgcls}
\end{table}

\noindent{\textbf{Effectiveness of ADTrans.}} From the results, we observe that our method can effectively improve the performance of baseline networks in nearly all metrics. Fig.~\ref{compare} presents a comparison between our method and plain PSGTR of the detailed recall@100 of SGDet task on part of predicate classes. ADTrans performs better than plain PSGTR on almost all the above predicate classes. When compared to IETrans \cite{zhang2022fine}, our method shows significant improvements in both recall and mean recall on all baseline models, indicating that our method can enhance the training dataset more effectively, avoiding noisy or redundant transfer processes. When it comes to PR, which takes into account both recall and mean recall, our method outperforms all the original models by significant margins. This suggests that our method not only improves the recall of the models but also balances the performance across different predicate labels, resulting in a more comprehensive evaluation of the models' performance.

\begin{table}[]
\tabcolsep=0.1cm

\begin{tabular}{l|cccccc}
\hline
\multirow{2}{*}{Method} & \multicolumn{6}{c}{Scene Graph Generation}                                                                         \\ \cline{2-7} 
                        & mR@20         & @50         & \multicolumn{1}{c|}{@100}        & F@20          & @50          & @100         \\ \hline
MOTIFS                  & 4.8           & 6.2           & \multicolumn{1}{c|}{7.1}           & 8.0           & 10.3          & 11.8          \\
\quad +ADTrans                & \textbf{10.6} & \textbf{15.5} & \multicolumn{1}{c|}{\textbf{18.1}} & \textbf{13.4} & \textbf{18.9} & \textbf{22.0} \\ \hline
VCTree                  & 5.2           & 6.7           & \multicolumn{1}{c|}{7.9}           & 8.7           & 11.0          & 13.0          \\
\quad +ADTrans                & \textbf{9.7}  & \textbf{12.5} & \multicolumn{1}{c|}{\textbf{16.9}} & \textbf{12.2} & \textbf{16.3} & \textbf{20.3} \\ \hline
GPSnet                  & 5.2           & 5.9           & \multicolumn{1}{c|}{7.1}           & 8.6           & 9.9           & 11.8          \\
\quad +ADTrans                & \textbf{12.3} & \textbf{15.8} & \multicolumn{1}{c|}{\textbf{19.2}} & \textbf{15.1} & \textbf{18.6} & \textbf{21.9} \\ \hline
\end{tabular}
\caption{The results (mR@K and F@K) on SGDET task of our method and other baselines on VG dataset.}
\label{sgdet}
\end{table}

\begin{table}[]

\tabcolsep=0.20cm
\begin{tabular}{ccc|cll}
\hline
\multicolumn{3}{c|}{Module} & \multicolumn{3}{c}{Scene Graph Generation}                                                   \\ \hline
IT      & PT      & RE      & R/mR@20                         & \multicolumn{1}{c}{R/mR@50} & \multicolumn{1}{c}{R/mR@100} \\ \hline
\XSolidBrush       & \XSolidBrush       & \XSolidBrush       & 28.4 / 16.6                     & 34.4 / 20.8                 & 36.3 / 22.1                  \\ \hline
\CheckmarkBold       & \XSolidBrush       & \XSolidBrush       & 26.2 / 24.9                     & 30.3 / 28.2                 & 30.7 / 29.2                  \\
\CheckmarkBold       & \CheckmarkBold       & \XSolidBrush       & 25.5 / 25.6                     & 29.2 / 29.1                 & 29.7 / 29.6                  \\
\CheckmarkBold       & \CheckmarkBold       & \CheckmarkBold       & \multicolumn{1}{l}{26.0 / 26.4} & 29.6 / 29.7                 & 30.0 / 30.0                  \\ \hline
\end{tabular}
\caption{Ablation study on model components. IT: Indistinguishable Triplet Transfer; PT: Potential Positive Transfer; RE: re-sampling.}
\label{component}
\end{table}

\noindent{\textbf{Expansibility of ADTrans.}} Applied with our ADTrans, the performances of all baseline models trained on all the datasets on all the tasks are greatly improved. With the observation that all the baseline models trained on the VG dataset have poorer performances compared to the same baseline models trained on the PSG dataset, VG dataset is more challenging for our ADTrans due to more biased annotations. Our method shows great expansibility on VG dataset, baseline models trained on the dataset achieve competitive performances on all the tasks.

\begin{figure*}[!t]
\centering   \includegraphics[width=\linewidth,height=5cm]{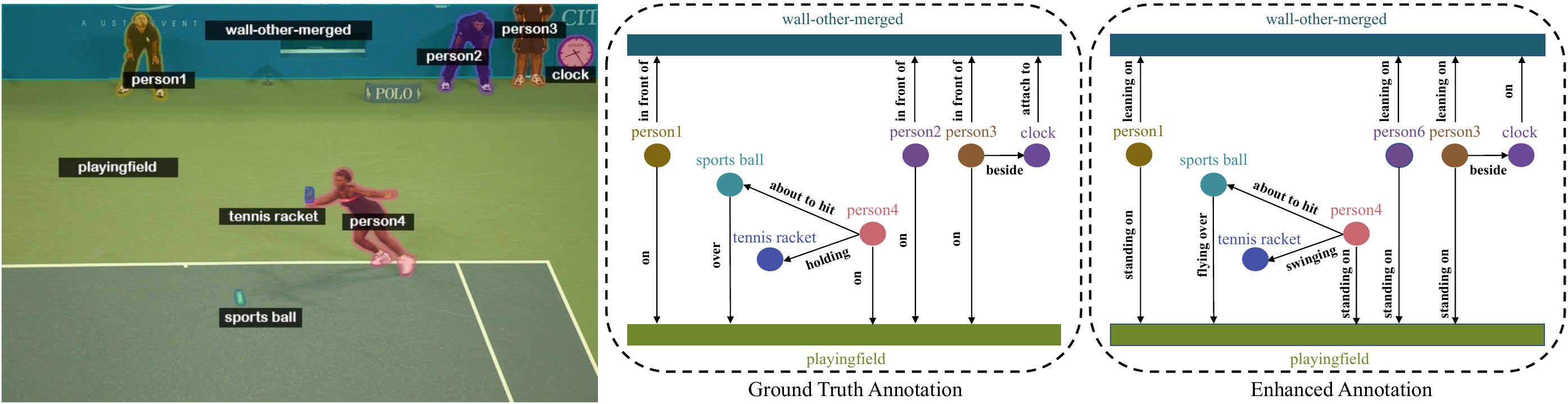}
    \caption{Visualization of the original dataset and a new dataset enhanced by our ADTRans. For the same image, we visualize its original biased annotations, and new annotations enhanced by our method. The enhanced dataset shows more informative annotations than the original one. These informative annotations promise the consistent training of models.}
    \label{fig:visualization}

\end{figure*}

\subsection{Ablation Studies} 
% In this part, we evaluate our method by applying different components, using different data processing methods during triplet transfer process, and using different contrastive learning methods.

\noindent\textbf{Different components in ADTrans framework.} We evaluate the importance of each component in our ADTrans. As shown in Table.~\ref{component}, we incrementally add one component to the plain baseline PSGTR \cite{yang2022psg} to validate their effectiveness.
The indistinguishable triplet transfer component provides the most promotion on performance. The reason is that, we transfer these inconsistent annotations to informative and consistent ones, so that models can learn a consistent mapping from visual to predicates.
The potentially positive transfer component additionally provides performance promotion. Our ADTrans transfers original NA samples, which are probably missed by annotators, to informative annotations. This step provides more reasonable training samples to construct a consistent training dataset.

\begin{table}[]

\tabcolsep=0.15cm

\begin{tabular}{c|ccc}
\hline
\multirow{2}{*}{Data Processing Method} & \multicolumn{3}{c}{SGDet}               \\ \cline{2-4} 
                                 & mR@20     & mR@50     & mR@100    \\ \hline
Original                         & 16.6 & 20.8 & 22.1 \\ \hline

Remove                           & 20.0 & 24.6 & 25.3 \\ \hline

Triplet Transfer                 & 24.9 & 28.2 & 29.2 \\ \hline
\end{tabular}
\caption{Ablation study on data processing methods. Triplet Transfer: Indistinguishable triplet transfer. Remove: simply removing all indistinguishable triplet pairs. Original: baseline method on the original dataset.}
\label{TAB:remove}
\end{table}

\noindent{\textbf{Different data processing methods during triplet transfer.}} To prove the effectiveness of our indistinguishable triplet transfer method and the harm of biased annotation, we design another simple method to process indistinguishable triplet pairs. Instead of adaptively transferring them to target pairs, we simply remove them from the original training dataset.
As shown in Table.~\ref{TAB:remove}, when comparing the simply removing method and the original baseline, the simply removing method surprisingly greatly overtakes the baseline model, indicating the serious conflict resulting from biased annotation within the original dataset. These biased annotated samples can not help the model during the training process, but will make it difficult for the model to distinguish each predicate label, resulting in a sharp decline in model performance. When comparing the triplet transfer method and the simply removing method, there are also great margins between these two methods. With the fact that our triplet transfer method greatly outperforms the simply removing method, we observe a promising textual information alignment process and the effectiveness of our adaptive triplet transfer method is proved.

% \begin{table}[]
% \tabcolsep=0.15cm
% \caption{Ablation study on sentence representation learning methods. BERT-ADTrans: BERT fine-tuned by our method. BERT-base: directly use original BERT model. Original: Plain PSGTR model without any representation method.}
% \label{bert}
% \begin{tabular}{c|ccc}
% \hline
% \multirow{2}{*}{Representation Learning} & \multicolumn{3}{c}{SGDet} \\ \cline{2-4} 
%                                          & mR@20  & mR@50  & mR@100  \\ \hline
% Original                                 & 16.6   & 20.8   & 22.1    \\ \hline
% BERT-base                                & 20.8   & 23.6   & 23.8    \\ \hline

% BERT-ADTrans                                  & 26.4   & 29.7   & 30.0    \\ \hline
% \end{tabular}
% \vspace{-1.0em}
% \end{table}

\begin{table}[]

\tabcolsep=0.05cm
\begin{tabular}{c|ccc}
\hline
\multirow{2}{*}{Contrastive Learning Method} & \multicolumn{3}{c}{SGDet} \\ \cline{2-4} 
                                         & mR@20  & mR@50  & mR@100  \\ \hline
Original                                 & 16.6   & 20.8   & 22.1    \\ \hline
InfoNCE                                & 21.0   & 25.8   & 26.3    \\ \hline

RCT                                  & 26.4   & 29.7   & 30.0    \\ \hline
\end{tabular}
\caption{Ablation study on contrastive learning methods. InfoNCE: the classic contrastive learning method. Original: the performance of original PSGTR. RCT: the proposed robust contrastive training method.}
\label{nce}
\end{table}

% \noindent{\textbf{Different sentence representation learning methods.}}  We evaluate the influence of different sentence representation learning methods in the indistinguishable triplet transfer process. As shown in Table.~\ref{bert}, we use BERT-ADTrans and BERT-base, respectively, to align the textual domain and the relationship domain during the data transfer process. We observe a great performance improvement even if only a BERT-base is used, indicating that the BERT-base controls the consistency during the data transfer process. Furthermore, we observe a huge promotion in model performance when our method is applied, especially on mR@100, with a 1.4-point increase,  meaning a more predicate-sensitive language model, indicating the effectiveness of our new sentence representation method.

\noindent{\textbf{Different contrastive learning methods.}}
As shown in Table.~\ref{nce}, though we observe improvement with the help of the InfoNCE, we can promote the model to a higher level with our method. An simply InfoNCE method increase the intra-class cohesion and inter-class separability, but cannot learn unbiased predicate representations because of biased annotations. As a result, biased annotations are inconspicuous and hard to be identified in a biased predicate representation embedding space. Our method utilizes invariant representation exploration and multistage data filtration to avoid the influence of biased annotations, and performs accurate biased-annotation identification.

\section{Conclusion}

We introduce a novel framework named ADTrans to alleviate the biased annotation problem in SGG. ADTrans transfers indistinguishable samples and potentially positive samples to promise a reasonable training dataset with more informative and standard-unified labels. Experiments demonstrate that ADTrans greatly enhances the models' performance on two datasets, achieving a new SOTA performance.

\section{Acknowledgements}
This research was supported in part by the National Key R\&D Program of China (2022ZD0119103), in part by the National Natural Science Foundation of China (Grant No. 62006123).
This research was supported by the MSIT (Ministry of Science, ICT), Korea,
under the ITRC (Information Technology Research Center) support program
(IITP-2023-2020-0-01789) supervised by the IITP (Institute for Information \&
Communications Technology Planning \& Evaluation).
\bibliography{aaai24}

\end{document}